# Reversible Image Watermarking for Health Informatics Systems Using Distortion Compensation in Wavelet Domain

Hamidreza Zarrabi, Mohsen Hajabdollahi, S.M.Reza Soroushmehr,
Nader Karimi, Shadrokh Samavi, Kayvan Najarian

*Abstract*— Reversible image watermarking guaranties restoration of both original cover and watermark logo from the watermarked image. Capacity and distortion of the image under reversible watermarking are two important parameters. In this study a reversible watermarking is investigated with focusing on increasing the embedding capacity and reducing the distortion in medical images. Integer wavelet transform is used for embedding where in each iteration, one watermark bit is embedded in one transform coefficient. We devise a novel approach that when a coefficient is modified in an iteration, the produced distortion is compensated in the next iteration. This distortion compensation method would result in low distortion rate. The proposed method is tested on four types of medical images including MRI of brain, cardiac MRI, MRI of breast, and intestinal polyp images. Using a one-level wavelet transform, maximum capacity of 1.5 BPP is obtained. Experimental results demonstrate that the proposed method is superior to the state-of-the-art works in terms of capacity and distortion.

## I. Introduction

Medical image analysis is extensively used to help physicians in medical applications and improves their diagnostic capabilities. Transmission of the medical information through a public network such as internet or computer network may be leaded to the security problems such as, modification and unauthorized access. In this regard, many researches were performed to address these problems and provided solutions for content authentication. Although digital image watermarking has addressed the aforementioned problems but medical image watermarking causes original image modification and distortion. Small distortion in medical images may have negative impact on physician diagnostic hence the original image content must be preserved. In recent years reversible watermarking has been introduced as an effective way to restore both the original image and watermark information. In this manner, physician diagnostic process and treatment are not affected, and patient privacies are kept. Reversible image watermarking can be considered as an essential part of health information system (HIS).

Recently, reversible watermarking has attracted a lot of attention in the research community. In many recent studies watermarking using transform domain, focused on the wavelet transform to reach better robustness. For example in [1], Haar discrete wavelet transform was used for reversible watermarking. In [2], difference expansion based reversible watermarking in Haar discrete wavelet domain was used. Kumar *et al.* employed singular value decomposition based reversible watermarking in discrete wavelet transform [3]. In [4], 4th level of discrete wavelet transform using a quantization function was used for embedding. Watermark information was encoded by BCH encoding to reach more security. Selvam *et al.* [5] firstly used integer wavelet transform to generate a transformed image and then they applied discrete gould transform in transformed coefficients and reached a capacity of 0.25 per pixel. In [6], Cohen-Daubechies-Fauraue integer wavelet transform was used. In order to prevent the overflow/underflow, histogram was firstly pre-processed, and then additional information was created. Companding was applied to compression and decompression of the transformed coefficients. Since in companding process, there was a probability of distortion, so, additional information was created.

Some of the previous studies have investigated the problem of the reversible watermarking with other transformation and techniques. In [7], mean value of transformed coefficients was utilized by slantlet for embedding. In order to increase the security, watermarking was transformed to Arnold domain and overflow/underflow was prevented by post-processing. In [8], initial using adaptive threshold detector algorithm, region of interest and non-region of interest was separated automatically. Then, embedding in each region separately was implemented using bin histogram. In [9], [10] and [11], reversible image watermarking based on Prediction Error Expansion was proposed. In [9], diagonal neighbors were considered as embedding locations, when overflow/underflow was occurred. In [12], histogram shifting was applied directly on pixels or prediction error, dynamically. In [13], two intelligent techniques including "Genetic algorithm" and "particle swarm optimization" were utilized for watermarking with interpolation error expansion. In [14], intermediate significant bit substitution was used as watermark embedding process. A fragile watermark was embedded for tamper detection. Watermark bits were encrypted before embedding in order to increase the security.

H.R Zarrabi, M. Hajabdollahi, and N. Karimi are with the Department of Electrical and Computer Engineering, Isfahan University of Technology, Isfahan 84156-83111, Iran.
S.M.R. Soroushmehr is with the Department of Computational Medicine and Bioinformatics and Michigan Center for Integrative Research in Critical Care, University of Michigan, Ann Arbor, MI, U.S.A.
S. Samavi is with the Department of Electrical and Computer Engineering, Isfahan University of Technology, Isfahan 84156-83111, Iran. He is also with the Department of Emergency Medicine, University of Michigan, Ann Arbor, MI, U.S.A.
K. Najarian is with the Department of Computational Medicine and Bioinformatics; Department of Emergency Medicine; and the Michigan Center for Integrative Research in Critical Care, University of Michigan, Ann Arbor, MI, U.S.A.

.

Reversible watermarking can be useful in case of encrypted images. Hence in [15], probabilistic properties of Paillier cryptosystem was used for encrypted images. In [16], block histogram shifting was used for encrypted image.

In this paper, a novel reversible watermarking method for medical images in telemedicine applications is proposed. Integer wavelet domain is utilized and watermark bits are embedded in each sub-band in two iterations. In the first iteration, coefficients are modified by embedding. In the second iteration, coefficients are modified in such a way that to be close to the original values. In this manner it is possible to have 2 bit embedding without any modifications. In other words, coefficient modification in the first iteration is compensated in the second iteration.

The rest of this paper is organized as follows. In Section II, proposed reversible watermarking based on integer wavelet transform is presented. Section III is dedicated to the experimental results. Finally in Section IV, our concluding remarks are presented.

## II. PROPOSED METHOD

In recent studies, discrete wavelet transform is extensively used as transformed domain based reversible image watermarking. In discrete wavelet transform, pixels are converted in form of integer values to the floating point one. Due to changing the coefficient values in embedding phase in form of truncation, preservation of original integer value cannot be guaranteed. To address this problem, in the proposed reversible image watermarking, integer to integer wavelet transform is exploited. The proposed method consists of two embedding and extraction phases. In the proposed algorithm, wavelet coefficients are converted to a binary map by:

$$Q(x) = mod\left(\left\lfloor \frac{x}{2} \right\rfloor, 2\right) \quad (1)$$

where "$mod$" is a function for calculating the remainder and $\lfloor . \rfloor$ is a floor function. According to the watermark bit and corresponding binary map, original coefficient value is changed by a constant value. This process may produce the same value for different coefficients and causes ambiguity during reconstruction of the original image. To avoid this ambiguity, tracker key is produced as side information and makes the algorithm reversible. In the followings, embedding and extraction phases are presented respectively.

### A. Embedding phase

An overview of the proposed embedding phase is presented in Fig. 1. System inputs are cover image and watermark logo and outputs are watermarked image and tracking key. At first, cover image is transformed by one-level integer wavelet transform and four wavelet sub-bands including LL, LH, HL and HH are calculated. LL sub-band has high sensitivity to human visual system and embedding in low frequency sub-band can be leaded to the perceptual distortion. So, according to the capacity requirements, three number of high frequency sab-bands (LH, HL, and HH) are selected for embedding. The number of watermark bits for embedding is divided in to the two equal parts for each selected embedding sub-band. Embedding phase has two iterations and in the first iteration, each bit of the first part is embedded in a coefficient. In the second iteration, second part of the watermark is embedded in the previously embedded coefficients.

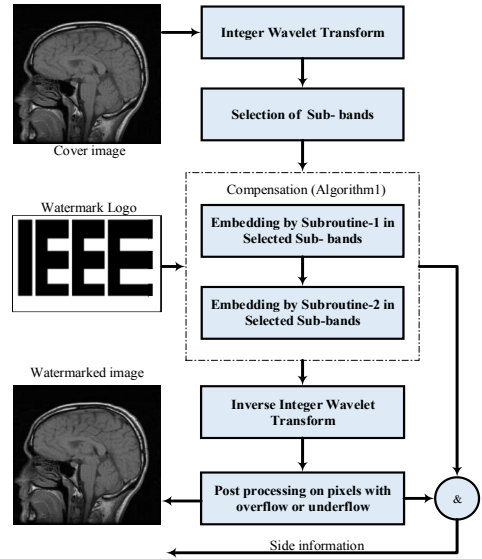

Figure 1. Block diagram of the embedding Procedure

Embedding algorithm for two iterations is explained in Algorithm1. Embedded image in wavelet domain is transformed back to the spatial domain with inverse integer wavelet transform. Embedding process may lead to a value outside of the acceptable range of the image values which are underflow in the case of pixels with value smaller than 0 and overflow in value greater than 255. For these cases, pixel values are truncated and their locations as well as their original values are considered as side information. Suppose that $c(u, v)$ is wavelet coefficient, $tkey(i)\ i = 1, ...$ is length of watermark as side information, $w$ is watermark, $c^w(u,v)$ is watermarked coefficient. Embedding algorithm (Algorithm1) is as follow.

| Algorithm1 : Embedding |
|---|
| **1:** Calculate $Q(c(u,v))$ by equation (1) |
| **2:**   **If $Q(c(u,v)) == 1$** |
| **3:**       $tkey(i) = 1$ |
| **4:**       In first iteration, apply subroutine 1 as follow |
| **5:**         If $w(i) = 0$ then $c^w(u,v) = c(u,v) + 2$ |
| **6:**         Else $c^w(u,v) = c(u,v)$ |
| **7:**       In second iteration, apply subroutine 2 as follow |
| **8:**         If $w(i) = 0$ then $c^w(u,v) = c(u,v) - 2$ |
| **9:**         Else $c^w(u,v) = c(u,v)$ |
| **10:**  **Else If $Q(c(u,v)) == 0$** |
| **11:**      $tkey(i) = 0$ |
| **12:**      In first iteration, apply subroutine 1 as follow |
| **13:**        If $w(i) = 1$ then $c^w(u,v) = c(u,v) + 2$ |
| **14:**        Else $c^w(u,v) = c(u,v)$ |
| **15:**      In second iteration, apply subroutine 2 as follow |
| **16:**        If $w(i) = 1$ then $c^w(u,v) = c(u,v) - 2$ |
| **17:**        Else $c^w(u,v) = c(u,v)$ |

Watermark information is embedded by iteration 2 in such a way that the variations due to iteration 1 are compensated. Embedding process has three advantages. First, the number of modified coefficients is small which can be leaded to the low perceptual distortion. Second, variation on coefficients duo to in iteration 1, can be compensated in iteration 2. Third, by the ability of embedding in each sub-band (LH, HL, HH), it is possible to increase the embedding capacity. Hence the maximum capacity is 1.5 BPP. Also by using simple methods such as duplicating, more robustness can be obtained.

*B. Extraction phase*

Overview of the proposed extraction phase is presented in Fig. 2. Inputs are watermarked image as well as tracking key and outputs are recovered original image and extracted watermark logo. At first, as a pre-processing stage we apply the side information including overflow or underflow and retrieve spatial domain version of the watermarked image. After the pre-processin, integer wavelet transform is applied on watermarked image and four sub-bands are obtained. Watermark information is extracted from the embedded sub-bands in two iterations by compensation algorithm. In each iteration, one bit is extracted from each wavelet coefficient. The order in which watermark information is extracted is inverse of the order in which it is embedded. So embedded watermark in the second iteration is extracted in the first iteration. Finally the original image is recovered by inverse integer wavelet transform. For extraction phase, suppose that $c^w(u,v)$ is coefficient value, $tkey(i)\ i = 1,...$ is length of watermark as side information, $w^e$ is as extracted watermark and $c^r(u,v)$ is recovered coefficient. The extraction algorithm (Algorithm2) of the proposed watermarking procedure is as follow.

**Algorithm 2: Extraction**

1: Calculate $Q(c^w(u,v))$ by equation (1)
2:     $w^e = Q(c^w(u,v))$
3:     **If** $tkey(i) = 1$ **then**
4:         In first iteration, apply subroutine 1 as follow
5:         If $(c^w(u,v))=0$ then $c^r(u,v) = c^w(u,v) + 2$
6:         Else $c^r(u,v) = c^w(u,v)$
7:         In second iteration , apply subroutine 2 as follow
8:         If $Q(c^w(u,v))=0$ then $c^r(u,v) = c^w(u,v) - 2$
9:         Else $c^r(u,v) = c^w(u,v)$
10:    **Else If** $tkey(i) = 0$ **then**
11:        In first iteration, apply subroutine 1 as follow
12:        If $Q(c^w(u,v))=1$ then $c^r(u,v) = c^w(u,v) + 2$
13:        Else $c^r(u,v) = c^w(u,v)$
14:        In second iteration, apply subroutine 2 as follow
15:        If $Q(c^w(u,v))=1$ then $c^r(u,v) = c^w(u,v) - 2$
16:        Else $c^r(u,v) = c^w(u,v)$

## III. EXPERIMENTAL RESULTS

Performance of the proposed method is tested on four grayscale medical image datasets including brain MRI [17], cardiac MRI [18], intestinal polyp [19] and breast MRI [20].

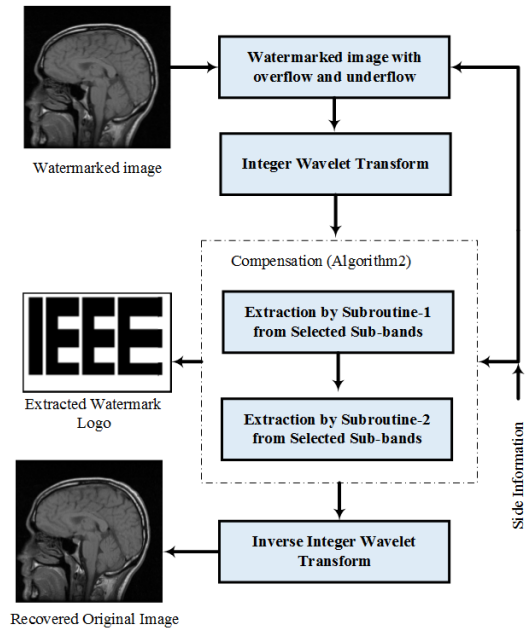

Figure 2. Block diagram of the extraction procedure.

Sample images from four medical datasets are illustrated in Fig. 3. All images have been resized to be 512×512. Input watermark is a binary image, consisting equal number of ones and zeros (49% ones and 51% zeros).

In Table 1, capacity-distortion results of the proposed method in case of four medical images are provided. Brain MRI, cardiac MRI, intestinal polyp and breast MRI are including 80, 70, 100 and 100 images which the average results of each dataset are as Table 1. Simulation results show the maximum capacity of 1.5 BPP with low distortion is obtained. Also it is observed from Table 1 that increasing the capacity from 0.1 to 1.5 BPP does not affect the visual quality significantly.

For comparison of the proposed method with other related methods, an experiment is performed in case of Lena image. A capacity-PSNR chart is illustrated in Fig. 4. It is observed that our method, for the same capacity, has better PSNR compared to the other six comparable methods. Also, visual quality after relatively high amount of embedding is acceptable. It is important to note that in the proposed algorithm, more capacity can be obtained by more transformation levels or iterations. Finally in order to evaluate the visual quality of the watermarked images, in Fig. 5 the original and the watermarked images are shown. In Fig. 5b, a sample watermark logo is shown for embedding. Fig. 5c shows watermarked image with capacity of 1.5 BPP. Fig. 5 shows that the watermarked images are not visually different from the original ones.

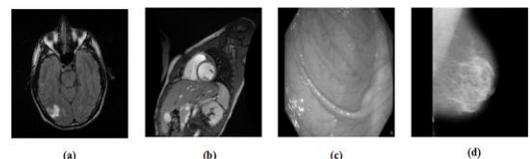

Figure .3 (a) Brain MRI, (b) Cardiac MRI, (c) Intestinal Polyp, (d) Breast MRI

TABLE 1. Average capacity (BPP) vs. distortion (PSNR) for four medical image datasets.

| Capacity (BPP) | PSNR(dB) | | | |
|---|---|---|---|---|
| | Brain MRI | Cardiac MRI | Intestinal Polyps | Breast MRI |
| 0.1 | 58.92 | 58.74 | 58.13 | 59.52 |
| 0.2 | 55.65 | 55.71 | 55.12 | 56.54 |
| 0.3 | 53.80 | 53.97 | 53.36 | 54.72 |
| 0.7 | 50.17 | 50.33 | 49.68 | 51.08 |
| 1 | 48.60 | 48.81 | 48.13 | 49.53 |
| 1.3 | 47.46 | 47.65 | 46.98 | 48.40 |
| 1.4 | 47.12 | 47.32 | 46.66 | 48.06 |
| 1.5 | 43.83 | 47.03 | 46.36 | 47.75 |

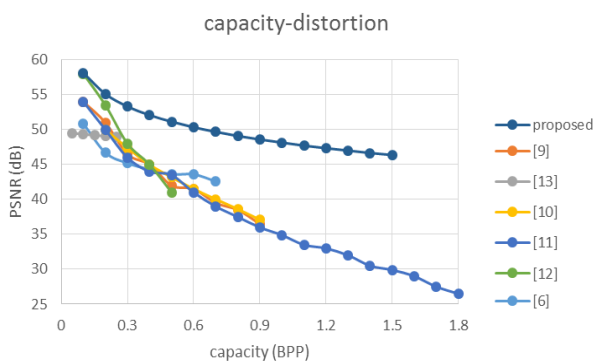

Figure 4. Comparison of capacity-distortion of proposed method with six other works for reversible embedding in Lena image.

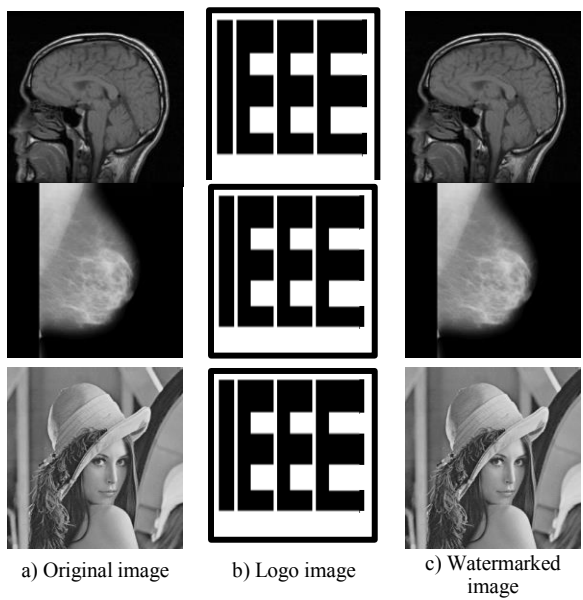

a) Original image    b) Logo image    c) Watermarked image

Figure 5. Visual quality of watermarked images using the proposed embedding method.

## IV. CONCLUSION

Novel reversible image watermarking method was presented in medical images based on integer wavelet transform. Improving distortion and embedding capacity was considered in the proposed method. Embedded process was performed in two iterations with a compensation method. It was possible for modified coefficients in the first iteration, to recover its original value in the second iteration. One bit was embedded on each iteration, hence maximum capacity of 1.5 BPP was obtained. Simulation results demonstrated that the proposed reversible image watermarking provided suitable capacity-distortion in comparison with the other methods.

## REFERENCES

[1] L.C. Huang, T.H. Feng, and M.S. Hwang, "A New Lossless Embedding Techniques Based on HDWT," *IETE Tech. Rev.*, vol. 34, no. 1, pp. 40–47, 2017.
[2] J. Park, S. Yu, and S. Kang, "Non-fragile High quality Reversible Watermarking for Compressed PNG image format using Haar Wavelet Transforms and Constraint Difference Expansions," *Int. J. Appl. Eng. Res.*, vol. 12, no. 5, pp. 582–590, 2017.
[3] M. Kumar, S. Agrawal, and T. Pant, "SVD-Based Fragile Reversible Data Hiding Using DWT," In *Proceedings of Fifth International Conference on Soft Computing for Problem Solving*, pp. 743-756, 2016.
[4] M. P. Turuk and A. P. Dhande, "A Novel Reversible Multiple Medical Image Watermarking for Health Information System," *J. Med. Syst.*, vol. 40, no. 12, 2016.
[5] P. Selvam., S. Balachandran, S. P. Iyer, and R Jayabal, "Hybrid Transform Based Reversible Watermarking Technique for Medical Images in Telemedicine Applications," *Opt. - Int. J. Light Electron Opt.*, pp. 655-671, 2017.
[6] M. Arsalan, A. S. Qureshi, A. Khan, and M. Rajarajan, "Protection of medical images and patient related information in healthcare: Using an intelligent and reversible watermarking technique," *Appl. Soft Comput. J.*, vol. 51, pp. 168–179, 2017.
[7] I. A. Ansari, M. Pant, and C. W. Ahn, "Artificial bee colony optimized robust-reversible image watermarking," *Multimed. Tools Appl.*, vol. 76, no. 17, pp. 18001–18025, 2017.
[8] Y. Yang, W. Zhang, D. Liang, and N. Yu, "A ROI-based high capacity reversible data hiding scheme with contrast enhancement for medical images," *Multimed. Tools Appl.*, pp. 1–23, 2017.
[9] I. Dragoi, and D. Coltuc, "Towards Overflow / Underflow Free PEE Reversible Watermarking," in *European Signal Processing Conference*, pp. 953–957, 2016.
[10] I. C. Dragoi and D. Coltuc, "Local-prediction-based difference expansion reversible watermarking," *IEEE Trans. Image Process.*, vol. 23, no. 4, pp. 1779–1790, 2014.
[11] X. Li, B. Yang, and T. Zeng, "Efficient Reversible Watermarking Based on Adaptive Prediction-Error Expansion and Pixel Selection," *IEEE Trans. IMAGE Process.*, vol. 20, no. 12, pp. 3524–3533, 2011.
[12] H. Y. Wu, "Reversible Watermarking Based on Invariant Image Classification and Dynamic Histogram Shifting," *Inf. Forensics Secur.*, vol. 8, no. 1, pp. 111–120, 2013.
[13] T. Naheed, I. Usman, T. M. Khan, A. H. Dar, and M. F. Shafique, "Intelligent reversible watermarking technique in medical images using GA and PSO," *Opt. - Int. J. Light Electron Opt.*, vol. 125, no. 11, pp. 2515–2525, 2014.
[14] S. A. Parah, F. Ahad, J. A. Sheikh, and G. M. Bhat, "Hiding clinical information in medical images: A new high capacity and reversible data hiding technique," *J. Biomed. Inform.*, vol. 66, pp. 214–230, 2017.
[15] S. Xiang and X. Luo, "Reversible Data Hiding in Homomorphic Encrypted Domain By Mirroring Ciphertext Group," *IEEE Trans. Circuits Syst. Video Technol.*, vol. 8215, no. c, pp. 1–12, 2017.
[16] Z. Yin *et al.*, "Reversible Data Hiding in Encrypted Image Based on Block Histogram Shifting," *ICASSP*, pp. 2129-2133, 2016.
[17] Brain MRI Images available at http://overcode.yak.net/15
[18] Cardiac MRI dataset available at http://www.cse.yorku.ca/~mridataset/
[19] ASU-Mayo Clinic Colonoscopy Video (c) Database available at https://polyp.grand-challenge.org/site/Polyp/AsuMayo/
[20] PEIPA, the Pilot European Image Processing Archive, available at http://peipa.essex.ac.uk/pix/mias